\documentclass{article}

\usepackage{arxiv}

\usepackage[utf8]{inputenc} % allow utf-8 input
\usepackage[T1]{fontenc}    % use 8-bit T1 fonts
\usepackage{hyperref}       % hyperlinks
\usepackage{url}            % simple URL typesetting
\usepackage{booktabs}       % professional-quality tables
\usepackage{amsfonts}       % blackboard math symbols
\usepackage{nicefrac}       % compact symbols for 1/2, etc.
\usepackage{microtype}      % microtypography
\usepackage{lipsum}		% Can be removed after putting your text content
\usepackage{graphicx}
\usepackage{natbib}
\usepackage{doi}
\usepackage{amsmath}

\title{Jacobian-Enforced Neural Networks (JENN) for Improved Data Assimilation Consistency in Dynamical Models}

\date{November 9, 2024}	% Here you can change the date presented in the paper title
\author{ \href{https://orcid.org/0000-0002-2832-4790}{\includegraphics[scale=0.06]{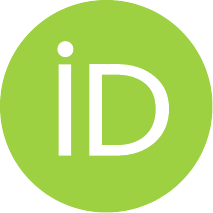}\hspace{1mm}Xiaoxu Tian}\thanks{ } \\ % footnote in this bracket
	Independent Researcher \\
	\texttt{xtian15@terpmail.umd.edu}
}

\date{}

\renewcommand{\shorttitle}{JENN for Data Assimilation}

\hypersetup{
pdftitle={JENN for Data Assimilation},
pdfsubject={JENN for Data Assimilation},
pdfauthor={Xiaoxu Tian},
pdfkeywords={Jacobian, Neural Network, Data Assimilation, Dynamical Models},
}

\begin{document}
\maketitle

\begin{abstract}
Machine learning-based weather models have shown great promise in producing accurate forecasts but have struggled when applied to data assimilation tasks, unlike traditional numerical weather prediction (NWP) models. This study introduces the Jacobian-Enforced Neural Network (JENN) framework, designed to enhance DA consistency in neural network (NN)-emulated dynamical systems. Using the Lorenz 96 model as an example, the approach demonstrates improved applicability of NNs in DA through explicit enforcement of Jacobian relationships. Training samples were generated from Lorenz 96 model forecasts, following a 1,000-model-time spin-up and subsequent collection of 80,000 data points over another 1,000 model times. The NN architecture includes an input layer of 40 neurons, two hidden layers with 256 units each employing hyperbolic tangent activation functions, and an output layer of 40 neurons without activation.

The JENN framework employs a two-step training process: an initial phase using standard prediction-label pairs to establish baseline forecast capability, followed by a secondary phase incorporating a customized loss function to enforce accurate Jacobian relationships. This loss function combines root mean square error (RMSE) between predicted and true state values with additional RMSE terms for tangent linear (TL) and adjoint (AD) emulation results, weighted to balance forecast accuracy and Jacobian sensitivity. To ensure consistency, the secondary training phase uses additional pairs of TL/AD inputs and labels calculated from the physical models. Notably, this approach does not require starting from scratch or structural modifications to the NN, making it readily applicable to pretrained models such as GraphCast, NeuralGCM, Pangu, or FuXi, facilitating their adaptation for DA tasks with minimal reconfiguration.

Experimental results demonstrate that the JENN framework preserves nonlinear forecast performance while significantly reducing noise in the TL and AD components, as well as in the overall Jacobian matrix. These findings highlight the potential of JENN to enhance the reliability of DA in advanced ML-based weather systems, paving the way for their seamless integration into operational forecasting workflows.

\end{abstract}

% keywords can be removed

\section{Introduction}
Accurate weather forecasting is essential for a wide range of applications, from agricultural planning to disaster preparedness. Traditional numerical weather prediction (NWP) models have long been the cornerstone of operational forecasting, relying on physical equations that govern atmospheric dynamics \citep{RN334}. In recent years, machine learning (ML) approaches, particularly neural networks, have shown great promise in emulating complex dynamical systems and producing accurate forecasts \citep{RN335, RN336}. These ML-based models offer the potential for faster computations and the ability to learn directly from observational data without explicit physical formulations \citep{RN313,RN326}.

However, despite their success in forecasting, ML models often face challenges when integrated into data assimilation (DA) systems \citep{RN307}. DA is a critical component of NWP, combining model forecasts with observational data to produce the best possible estimate of the current state of the atmosphere \citep{RN52}. Effective DA relies on an accurate representation of the model sensitivity to initial conditions, typically quantified through the model’s Jacobian or tangent linear (TL) and adjoint (AD) models \citep{RN113}. Traditional NWP models inherently possess well-defined TL and AD structures due to their explicit physical formulations, enabling efficient and reliable DA processes.

In contrast, neural networks lack explicit representations of their internal sensitivities, making it difficult to derive accurate TL and AD models necessary for DA \citep{RN307,RN338}. This limitation hampers the integration of ML models into operational DA frameworks, restricting their applicability to real-world forecasting systems. Addressing this gap is essential for leveraging the full potential of ML models in meteorology.

This study introduces the Jacobian-Enforced Neural Network (JENN), a novel approach designed to enhance the data assimilation consistency of neural network-based dynamical models. By incorporating Jacobian enforcement directly into the training process, JENN aims to improve the alignment of the internal sensitivities of the neural network with those of the true dynamical system. The Lorenz 96 model \citep{RN337}, a simplified representation of atmospheric dynamics known for its chaotic behavior, is used as a testbed to demonstrate the effectiveness of the proposed method.

The key contributions of this work are:
1.	Development of the JENN Framework: We propose a training methodology that enforces Jacobian consistency by incorporating tangent linear information into the loss function. This approach balances forecast accuracy with accurate representation of model sensitivities.
2. Improved data assimilation consistency: Through experiments with the Lorenz 96 model, we demonstrate that JENN preserves nonlinear forecast performance while significantly improving the quality of the neural network’s TL and AD representations.
3. Potential for operational integration: The findings suggest that JENN can facilitate the integration of ML-based models into existing DA systems, paving the way for more efficient and accurate weather forecasting.

The remainder of this paper is organized as follows. Section 2 details the methodology, including the neural network architecture and the Jacobian enforcement strategy. Section 3 presents the experimental results and analyzes the performance of JENN compared to standard neural networks. Finally, Section 4 discusses the implications of the findings and outlines the directions for future research.

\section{Methodology}

\subsection{Lorenz96 Model and Its Neural Network Emulator}
The Lorenz96 model is a widely used conceptual model in atmospheric sciences for studying chaotic systems and testing data assimilation techniques. Introduced by Edward Lorenz \citep{RN337}, it captures essential features of atmospheric dynamics such as nonlinearity and sensitivity to initial conditions, making it an ideal testbed for numerical experiments.

The Lorenz96 model is defined by a set of N coupled ordinary differential equations:

\begin{equation}
\frac{dx_i}{dt} = (x_{i+1} - x_{i-2}) x_{i-1} - x_i + F
\label{eq:lorenz96}
\end{equation}

where $x_i$ represents the state of the system at the $i$-th grid point and $F$ is a constant forcing term. The indices are cyclic, so $x_{-1}=x_{N-1}$ and $x_{N+1}=x_1$. 

In this study, the parameters are set as $N=40$ and $F=8$ as commonly used to induce chaotic behavior that resembles atmospheric dynamics at a simplified level.

The goal of the neural network emulator is to approximate the dynamics of the Lorenz96 model by learning the mapping from the current state $\mathbf{x}(t)$ to the next state $\mathbf{x}(t+\Delta t)$. The architecture of the neural network is as follows:

\begin{description}
  \item[$\cdot$ Input layer] An input vector $\mathbf{x}(t)$, representing the current state of the system.
  \item[$\cdot$ Hidden layers] Two fully connected hidden layers, each with 256 neurons. The activation function used is the hyperbolic tangent function ($\tanh$), which introduces nonlinearity and allows the network to model complex relationships.
  \item[$\cdot$ Output layer] A fully connected layer with 40 neurons, producing the predicted next state $\mathbf{x}(t+\Delta t)$. In this layer, no activation function is applied to allow the network to output any real-valued numbers.
\end{description}

An illustration of the NN structure can be seen in Figure 1. To generate data for training the neural network emulator, the Lorenz96 model is numerically integrated using a fourth-order Runge-Kutta scheme with a time step $\Delta t=0.0125$. Before any actual training samples are collected, the model will spin up for 1000 model time units (equivalent to 80,000 time steps) to allow transient behaviors to dissipate, ensuring that the system\'s trajectory lies on the attractor. After the spin-up period, the model is further integrated for 1,000 model time units (80,000 time steps). At each time step $t$, we record the current state $\mathbf{x}(t)$ and the next state $\mathbf{x}(t+\Delta t)$, forming input-target pairs for training. This process yields a dataset of two million samples, capturing the dynamics of the Lorenz96 system over a substantial period. The training of the neural network emulator involves minimizing the discrepancy between the predicted next state $\mathbf{x}(t)$ and the true next state $\mathbf{x}(t+\Delta t)$. The standard loss function used is the root-mean-square error (RMSE):

\begin{equation}
\mathcal{L}_{\text{forecast}} = \sqrt{\frac{1}{N} \sum{i=1}^{N} \left( x_i(t + \Delta t) - x_{i,true}(t + \Delta t) \right)^2 }.
\label{eq:forecast_loss}
\end{equation}

This loss function measures the network\'s ability to replicate the Lorenz96 dynamics over one time step.

\subsection{Jacobian Enforcement for Data Assimilation}
In data assimilation, it is crucial that the network not only predicts the next state accurately but also captures the sensitivities of the system to perturbations in the initial conditions. This sensitivity is represented by the Jacobian matrix $J$, whose elements are defined as:

\begin{equation}
J_{ij} = \frac{\partial x_i(t + \Delta t)}{\partial x_j(t)}, \quad i, j = 1, 2, \dots, N.
\end{equation}

To enforce Jacobian, two sets of perturbations are used as the inputs for the Lorenz 96 tangent linear ($\delta \mathbf{x}(t)$) and adjoint ($\hat{\mathbf{x}}(t+\Delta t)$) models. In both cases, the perturbations $\delta \mathbf{x}(t)$ are scaled to approximately 1\% of the corresponding state values. The true changes in the next state due to these perturbations $\delta \mathbf{x}(t+\Delta t)$ and $\hat{\mathbf{x}}(t)$ are computed using the Lorenz96 model’s tangent linear and adjoint model. A custom loss function combines the forecast loss and a Jacobian enforcement term:

\begin{equation}
\mathcal{L}_{\text{total}} = \alpha \mathcal{L}_{\text{forecast}} + \beta \mathcal{L}_{\text{TLM}} + \gamma
\mathcal{L}_{\text{ADJ}},
\label{eq:total_loss}
\end{equation}

where $\mathcal{L}_{\text{TLM}}$ is the RMSE between the neural network’s predicted perturbations $\delta \mathbf{x}(t+\Delta t)$ and the true perturbations $\delta \mathbf{x}_{true}(t+\Delta t)$ and $\mathcal{L}_{\text{ADJ}}$ RMSE between the neural network's adjoint $\hat{\mathbf{x}}(t)$ and the true adjoint $\hat{\mathbf{x}}_{true}(t)$. The $\alpha$, $\beta$, and $\gamma$ are weighting coefficients that balance the importance of forecast accuracy and Jacobian consistency. In other words, by adding an additional $\mathcal{L}_{\text{TLM}}$ and $\mathcal{L}_{\text{ADJ}}$ in the loss function, the model is trained to make accurate predictions in the space of state vector $\mathbf{x}$ and their perturbations $\delta \mathbf{x}$ following the Jacobian of the model dynamics. A schematic illustration of the components included in the loss function can be found in Figure 1. \\

In the training implementation, the neural network is initially trained using only the forecast loss ($\mathcal{L}_{\text{forecast}}$) in equation (2) with the L-BFGS optimization algorithm until convergence, where L-BFGS identifies a minimum value. Once the forecast model has converged, the network undergoes a second training phase using the total loss ($\mathcal{L}_{\text{total}}$) in equation (4). To compute $\mathcal{L}_{\text{TLM}}$ and $\mathcal{L}_{\text{ADJ}}$, random perturbations are applied at randomly selected locations to the TL/AD components of the pretrained neural network. The resulting responses are then compared to those from the TL/AD models of the physical Lorenz 96 model to calculate the RMSE. During this phase, the trained weights and biases are updated and saved when L-BFGS reaches convergence. This two-step approach enforces Jacobian consistency without compromising the already established forecasting capability, ensuring the neural network retains its predictive accuracy while enhancing its internal representation of system sensitivities. This improvement makes the model more suitable for integration into data assimilation systems.

\section{Results and Discussion}

The two-step training approach aims at improving the neural network’s ability to emulate tangent linear perturbations and adjoint sensitivities while preserving the accuracy of nonlinear forecast dynamics. By initially focusing on the forecast loss ($\mathcal{L}_{\text{forecast}}$), the model builds a robust foundation for accurate state predictions. The subsequent refinement using the total loss ($\mathcal{L}_{\text{total}}$) enforces Jacobian consistency, ensuring that the model’s internal representations align with the physical system’s sensitivities. Importantly, the accuracy of the nonlinear forecast is preserved throughout the training process. As shown in Figure 2, the JENN predictions ($\mathbf{y}_{JENN}$) closely match the true forecast ($\mathbf{y}_{true}$), with minimal deviations compared to the standard neural network ($\mathbf{y}_{NN}$). The three curves in Figure 2a are so closely aligned that they almost entirely overlap. To better illustrate the differences between the emulated results and the true $\mathbf{y}$, the absolute differences are plotted in Figure 2b. The comparison reveals that the magnitudes of $\mathbf{y}_{JENN}$ relative to $\mathbf{y}_{true}$ are comparable to those of $\mathbf{y}_{NN}$, indicating no degradation in forecast accuracy after the second training phase.

The tangent linear responses exhibit notable improvements in accuracy following the JENN training. Figure 3 demonstrates that the TLM predictions from the JENN framework ($\delta \mathbf{y}_{JENN}$) are significantly closer to the true TLM ($\delta \mathbf{y}_{true}$) compared to the standard neural network ($\delta \mathbf{y}_{NN}$). In Figure 3a, the results from both the standard NN and the JENN framework align well with the true perturbation responses. However, the absolute differences shown in Figure 3b reveal that the JENN framework significantly reduces noise and unphysical oscillations, particularly in regions with sharp gradients. This improvement highlights the ability of JENN to accurately capture the system’s sensitivity to perturbations, a critical requirement for data assimilation applications.

The adjoint responses further underscore the benefits of Jacobian enforcement. As shown in Figure 4, the adjoint predictions from the JENN framework ($\hat{\mathbf{y}}_{JENN}$) exhibit smaller deviations from the true adjoint ($\hat{\mathbf{y}}_{true}$) compared to those from the standard neural network ($\hat{\mathbf{y}}_{NN}$). The absolute differences presented in Figure 4b emphasize the substantial improvements achieved in adjoint consistency, particularly in accurately capturing the sensitivities of the input state to the forecast output. This enhancement ensures that the neural network is better equipped to provide accurate gradient information, making it more suitable for variational data assimilation frameworks.

The heat map of Jacobian matrix comparisons in Figure 5 offer a detailed perspective on the improvements in overall Jacobian accuracy, which is critical for data assimilation methods. The JENN framework significantly reduces the deviations between the learned Jacobian ($J_{JENN}$) and the true Jacobian ($J_{true}$), as evident in the bottom right panel in Fig. 5. In contrast, the standard neural network displays larger deviations, with notable noise and inaccuracies across the matrix. These findings highlight the ability of the JENN framework to substantially improve the model’s representation of system dynamics without requiring explicit access to the full Jacobian during training, relying instead on the tangent linear and adjoint information. The improvements in TLM and ADJ consistency, as well as the reduction in Jacobian deviations, make the JENN framework a promising candidate for integration into data assimilation systems. Accurate representation of sensitivities is crucial for methods such as 4DVar and ensemble-based DA, where the quality of the tangent linear and adjoint models directly impacts the analysis accuracy. By preserving forecast performance while enhancing sensitivity representations, JENN fills the gap between traditional NWP models and ML-based weather models, offering a scalable and effective solution for operational applications.

\section{Summary and Conclusion}

This study presents a Jacobian-Enforced Neural Network (JENN) framework designed to enhance the accuracy and physical consistency of tangent linear and adjoint models in machine learning-based weather systems. By applying a two-step training approach, JENN ensures improved representation of system sensitivities while maintaining nonlinear forecast accuracy. The training process involves an initial phase focused solely on forecast loss ($\mathcal{L}_{\text{forecast}}$) to establish a solid foundation for accurate predictions, followed by a refinement phase that incorporates tangent linear and adjoint loss terms ($\mathcal{L}_{\text{TLM}}$ and $\mathcal{L}_{\text{ADJ}}$) to enforce Jacobian consistency.

The results demonstrate significant improvements in both tangent linear and adjoint responses under the JENN framework. Tangent linear predictions ($\delta \mathbf{y}_{JENN}$) exhibit reduced noise and improved alignment with the true system sensitivities ($\delta \mathbf{y}_{true}$), as shown in Figure 3. Similarly, adjoint predictions ($\hat{\mathbf{y}}_{JENN}$) are notably closer to the true adjoint ($\hat{\mathbf{y}}_{true}$), with substantial reductions in deviations, particularly in regions of high sensitivity (Figure 4). The heat map comparisons of the Jacobians (Figure 5) further highlight the effectiveness of JENN in improving the overall dynamical consistency of the learned model. The framework achieves these enhancements without requiring explicit access to the full Jacobian during training, relying instead on tangent linear and adjoint computations.

The improvements achieved through JENN have important implications for data assimilation systems, particularly those requiring accurate sensitivity information, such as four-dimensional variational (4DVar) assimilation. By filling the gap between traditional numerical weather prediction models and machine learning-based systems, JENN offers a scalable solution that enhances the suitability of neural networks for operational forecasting and data assimilation applications.

Future work will focus on extending the JENN framework to more complex atmospheric models and higher-dimensional systems. Additional investigations will explore the impact of neural network architecture, training sample size, and hyperparameter tuning to further optimize performance. The integration of JENN into operational frameworks has the potential to transform the use of machine learning in weather prediction, offering new opportunities for improved forecasting and data assimilation in increasingly sophisticated modeling environments.

\bibliographystyle{unsrtnat}
\bibliography{references}  %%% Uncomment this line and comment out the ``thebibliography'' section below to use the external .bib file (using bibtex) .

\newpage
\begin{figure}
\noindent\includegraphics[width=\textwidth]{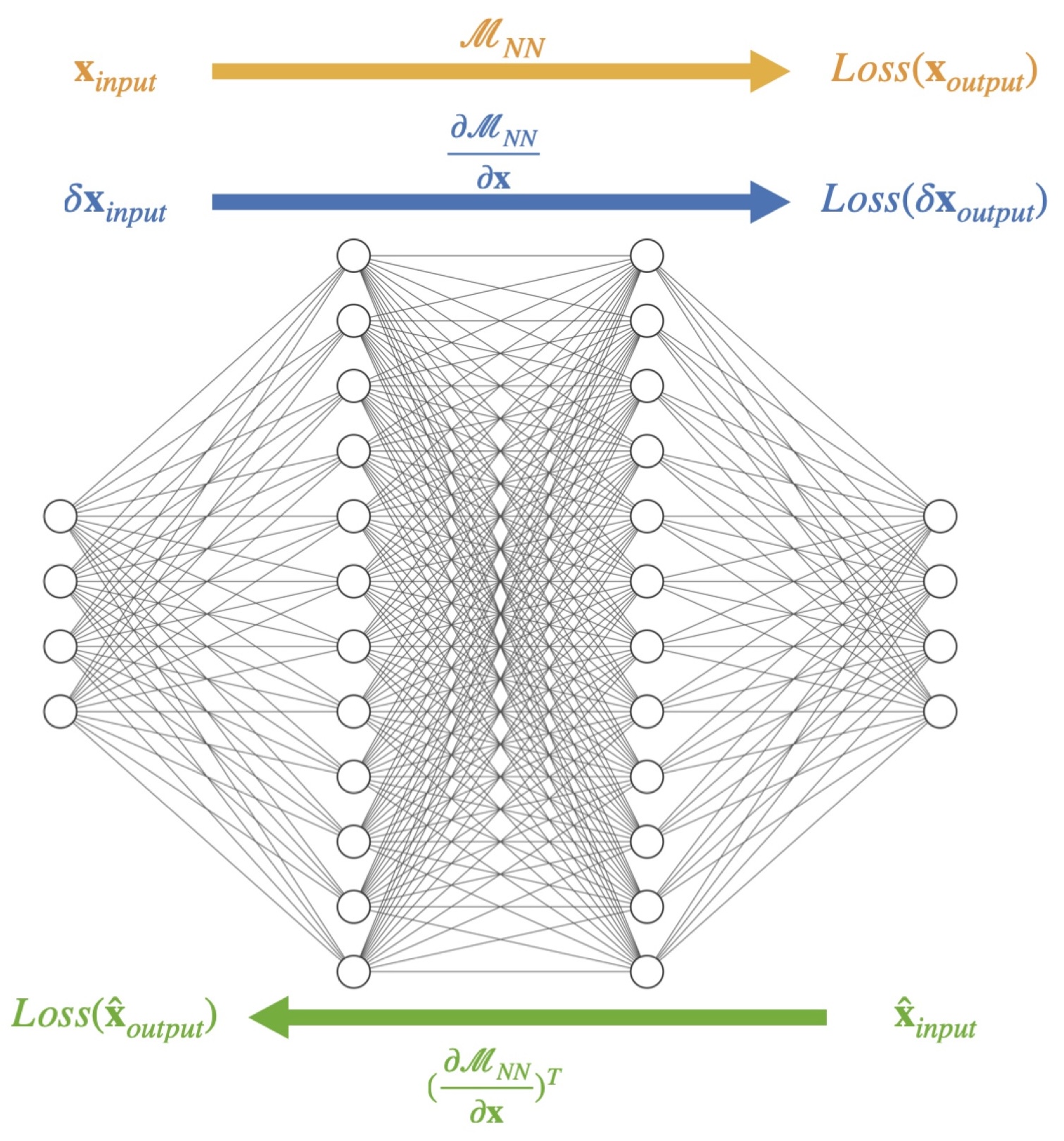}
\caption{
An illustration of neural network structure to emulate the Lorenz 96 model, with an input layer of 40 nodes, two hidden layers both of 256 nodes, and an output layer of 40 nodes. The diagram highlights three key data flows: the nonlinear forward pass (orange), the tangent linear propagation (blue), and the backward adjoint propagation (green). Each stream contributes to the total loss function, combining nonlinear forecast loss, tangent linear loss, and adjoint loss.
}
%\label{pngfiguresample}
\end{figure}

\newpage
\begin{figure}
\noindent\includegraphics[width=\textwidth]{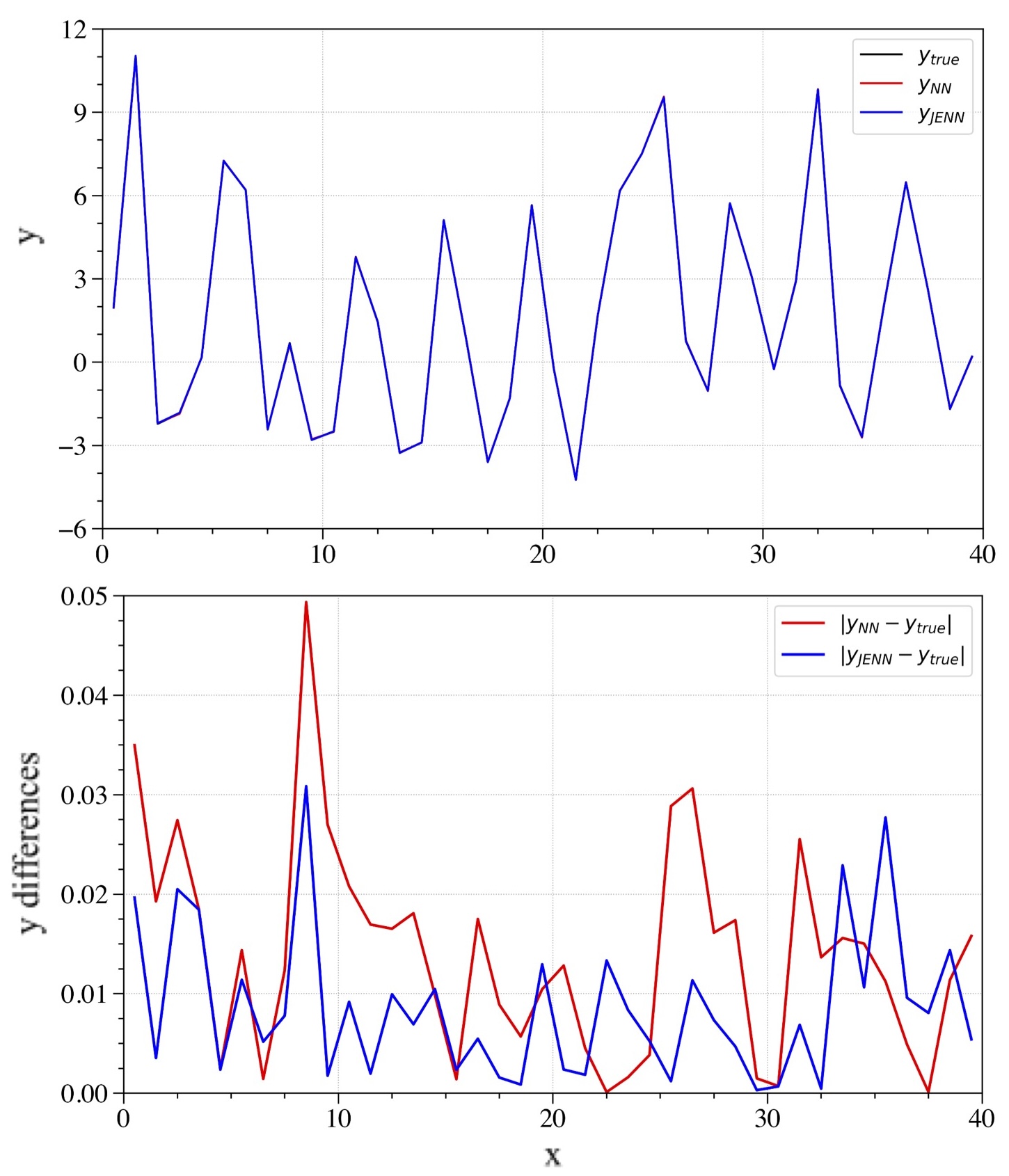}
\caption{
An illustration of neural network structure to emulate the Lorenz 96 model, with an input layer of 40 nodes, two hidden layers both of 256 nodes, and an output layer of 40 nodes. The diagram highlights three key data flows: the nonlinear forward pass (orange), the tangent linear propagation (blue), and the backward adjoint propagation (green). Each stream contributes to the total loss function, combining nonlinear forecast loss, tangent linear loss, and adjoint loss.
}
%\label{pngfiguresample}
\end{figure}

\newpage
\begin{figure}
\noindent\includegraphics[width=\textwidth]{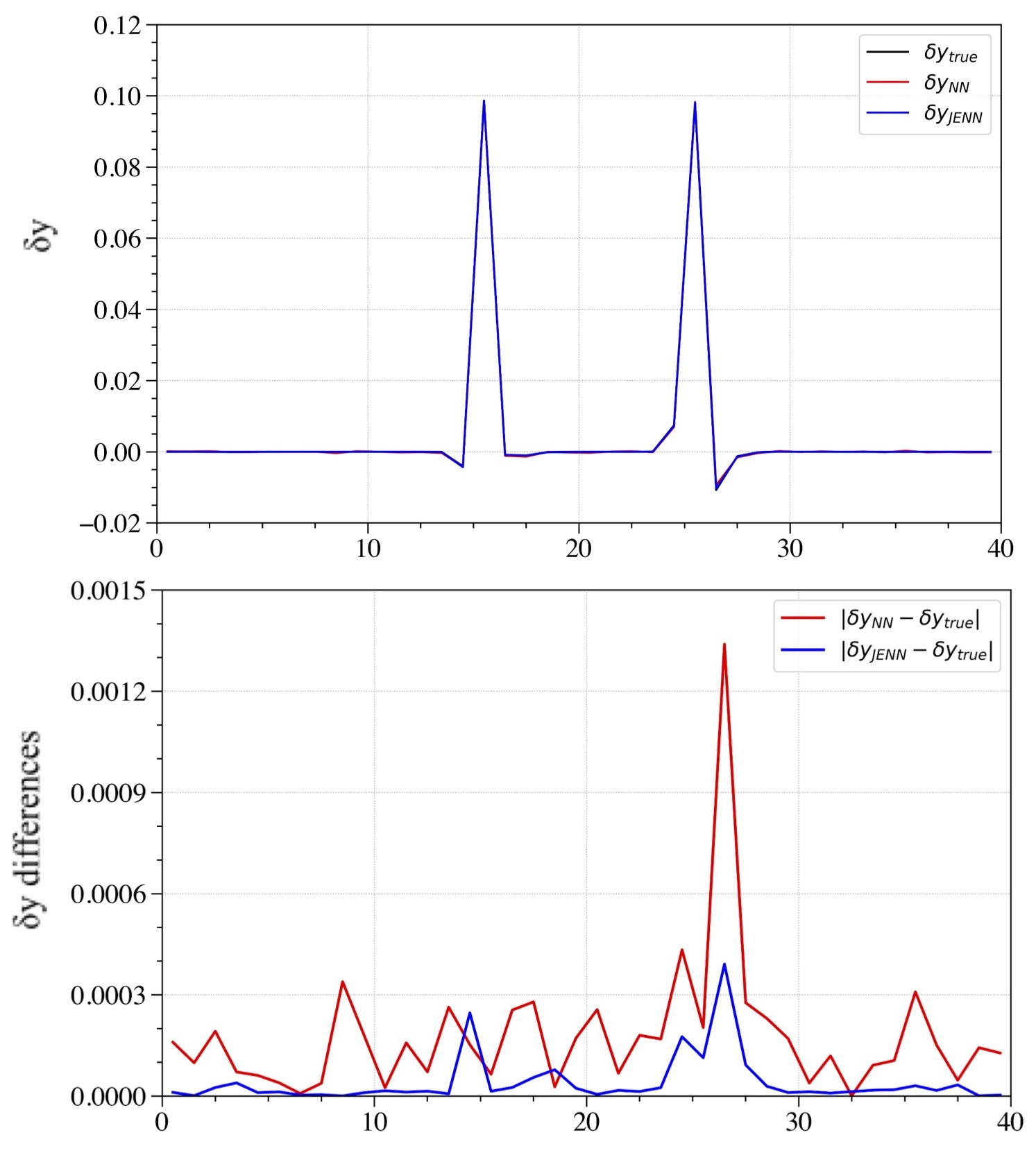}
\caption{
An illustration of neural network structure to emulate the Lorenz 96 model, with an input layer of 40 nodes, two hidden layers both of 256 nodes, and an output layer of 40 nodes. The diagram highlights three key data flows: the nonlinear forward pass (orange), the tangent linear propagation (blue), and the backward adjoint propagation (green). Each stream contributes to the total loss function, combining nonlinear forecast loss, tangent linear loss, and adjoint loss.
}
%\label{pngfiguresample}
\end{figure}

\newpage
\begin{figure}
\noindent\includegraphics[width=\textwidth]{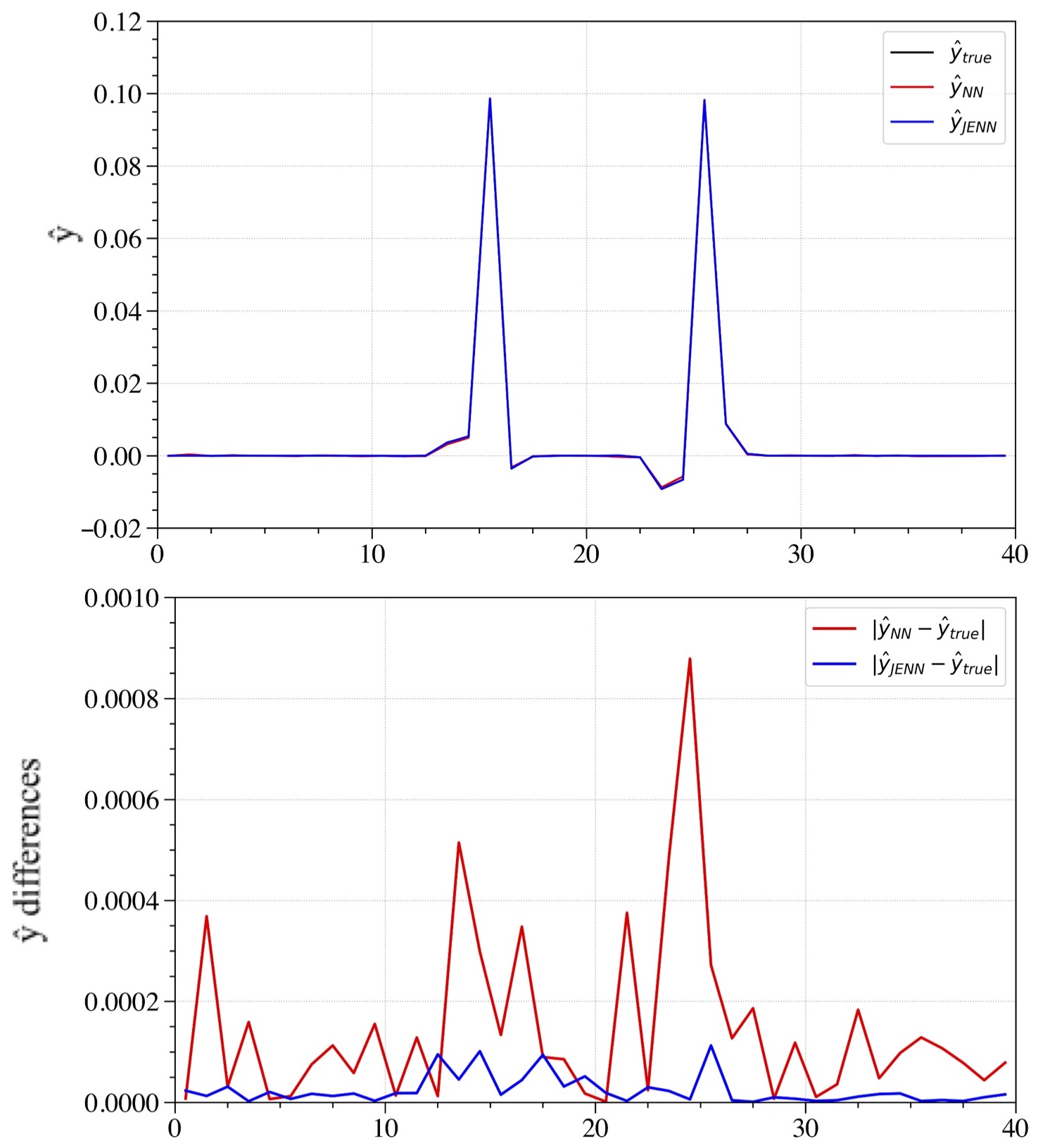}
\caption{
An illustration of neural network structure to emulate the Lorenz 96 model, with an input layer of 40 nodes, two hidden layers both of 256 nodes, and an output layer of 40 nodes. The diagram highlights three key data flows: the nonlinear forward pass (orange), the tangent linear propagation (blue), and the backward adjoint propagation (green). Each stream contributes to the total loss function, combining nonlinear forecast loss, tangent linear loss, and adjoint loss.
}
%\label{pngfiguresample}
\end{figure}

\newpage
\begin{figure}
\noindent\includegraphics[width=\textwidth]{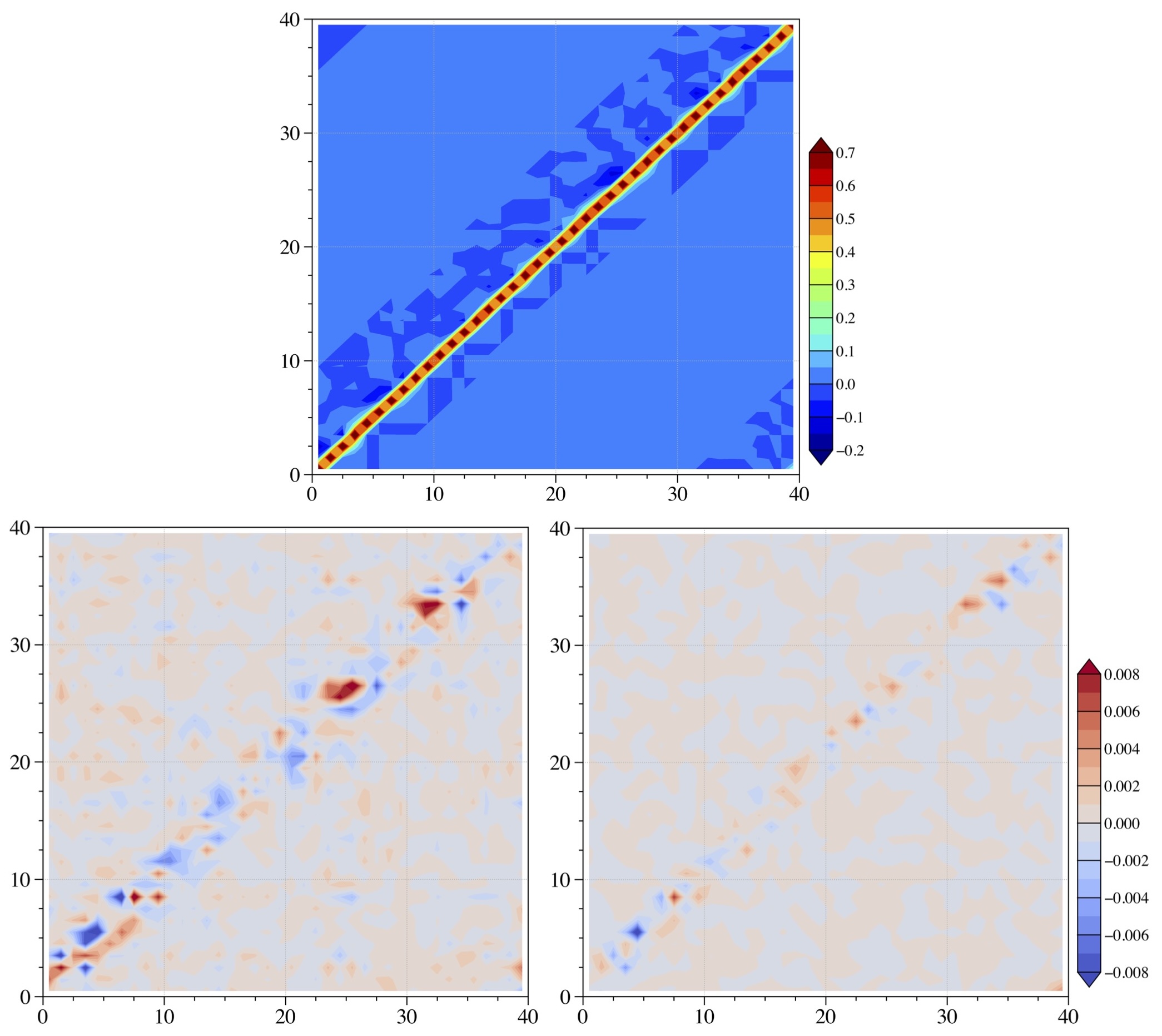}
\caption{
An illustration of neural network structure to emulate the Lorenz 96 model, with an input layer of 40 nodes, two hidden layers both of 256 nodes, and an output layer of 40 nodes. The diagram highlights three key data flows: the nonlinear forward pass (orange), the tangent linear propagation (blue), and the backward adjoint propagation (green). Each stream contributes to the total loss function, combining nonlinear forecast loss, tangent linear loss, and adjoint loss.
}
%\label{pngfiguresample}
\end{figure}





% \end{thebibliography}

\end{document}